\documentclass{article}

\usepackage{microtype}
\usepackage{graphicx}
\usepackage{subcaption}
\usepackage{booktabs} 

\usepackage{mathtools}
\usepackage{dsfont}
\usepackage{amsfonts}
\usepackage[version=4]{mhchem}
\usepackage{textcomp} 

\usepackage{multirow} 
\usepackage{enumitem}
\setlist[itemize]{itemsep=0pt, topsep=0pt}

\newcommand*{\tran}{^{\mkern-1.5mu\mathsf{T}}}

\usepackage{hyperref}

\hypersetup{
  pdflang={en-US}, 
}

\usepackage[accepted]{icml2020}

\icmltitlerunning{}

\begin{document}

\twocolumn[
\icmltitle{Reinforcement Learning for Molecular Design \\ Guided by Quantum Mechanics}



\icmlsetsymbol{equal}{*}

\begin{icmlauthorlist}
\icmlauthor{Gregor N.\ C.\ Simm}{equal,cam}
\icmlauthor{Robert Pinsler}{equal,cam}
\icmlauthor{Jos\'e Miguel Hern\'andez-Lobato}{cam}
\end{icmlauthorlist}

\icmlaffiliation{cam}{Department of Engineering, University of Cambridge, Cambridge, UK}

\icmlcorrespondingauthor{Gregor N.\ C.\ Simm}{gncs2@cam.ac.uk}

\icmlkeywords{Machine Learning, ICML}

\vskip 0.3in
]



\printAffiliationsAndNotice{\icmlEqualContribution} 

\begin{abstract}
Automating molecular design using deep reinforcement learning (RL) holds the promise of accelerating the discovery of new chemical compounds.
Existing approaches work with molecular graphs and thus ignore the location of atoms in space, which restricts them to 1) generating single organic molecules and 2) heuristic reward functions.
To address this, we present a novel RL formulation for molecular design in Cartesian coordinates, thereby extending the class of molecules that can be built.
Our reward function is directly based on fundamental physical properties such as the energy, which we approximate via fast quantum-chemical methods.
To enable progress towards de-novo molecular design, we introduce \mbox{\textsc{MolGym}},
an RL environment comprising several challenging molecular design tasks along with baselines.
In our experiments, we show that our agent can efficiently learn to solve these tasks from scratch by working in a translation and rotation invariant state-action space.
\end{abstract}

\section{Introduction}

Finding new chemical compounds with desired properties is a challenging task with important applications such as \emph{de novo} drug design and materials discovery \citep{Schneider2019}.
The diversity of synthetically feasible chemicals that can be considered as potential drug-like molecules was estimated to be between $10^{30}$ and $10^{60}$ \citep{Polishchuk2013}, making exhaustive search hopeless.

Recent applications of machine learning have accelerated the search for new molecules with specific desired properties.
Generative models such as variational autoencoders (VAEs) \citep{Gomez-Bombarelli2016}, recurrent neural networks (RNNs) \citep{Segler2018}, and generative adversarial networks (GANs) \citep{DeCao2018} have been successfully applied to propose potential drug candidates.
Despite recent advances in generating valid structures, proposing truly novel molecules beyond the training data distribution remains a challenging task.
This issue is exacerbated for many classes of molecules (e.g. transition metals), where such a representative dataset is not even available.

\begin{figure}[t]
\centering
\includegraphics{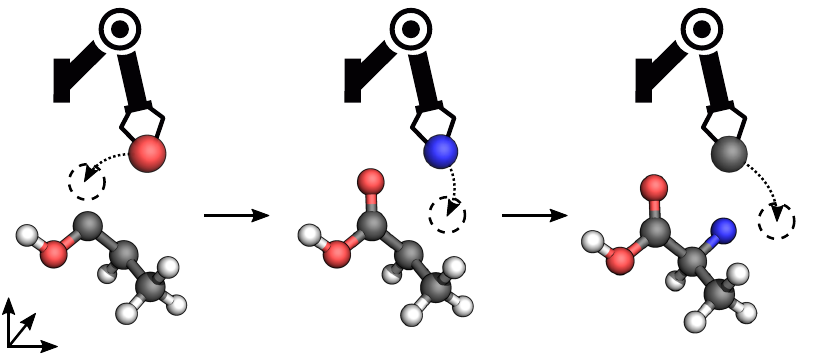}
\caption{
Visualization of the molecular design process presented in this work.
The RL agent (depicted by a robot arm) sequentially places atoms onto a canvas.
By working directly in Cartesian coordinates, the agent learns to build structures from a very general class of molecules.
Learning is guided by a reward that encodes fundamental physical properties.
Bonds are only for illustration.
}
\label{fig:intro}
\end{figure}

An alternative strategy is to employ RL, in which an agent builds new molecules in a step-wise fashion (e.g., \citet{Olivecrona2017}, \citet{Guimaraes2018}, \citet{Zhou2019}, \citet{Zhavoronkov2019}).
Training an RL agent only requires samples from a reward function, alleviating the need for an existing dataset of molecules.
However, the choice of state representation in current models still severely limits the class of molecules that can be generated.
In particular, molecules are commonly described by graphs, where atoms and bonds are represented by nodes and edges, respectively.
Since a graph is a simplified model of the physical representation of molecules in the real world, one is limited to the generation of single organic molecules. 
Other types of molecules cannot be appropriately described as this representation lacks important three-dimensional (3D) information, i.e. the relative position of atoms in space.
For example, systems consisting of multiple molecules cannot be generated for this reason.
Furthermore, it prohibits the use of reward functions based on fundamental physical laws; instead, one has to resort to heuristic physicochemical parameters, e.g. the Wildman-Crippen partition coefficient \citep{Wildman1999}.
Lastly, it is not possible to impose geometric constraints on the design process, e.g. those given by the binding pocket of a protein which the generated molecule is supposed to target.

In this work, we introduce a novel RL formulation for molecular design in which an agent places atoms from a given bag of atoms onto a 3D canvas (see Fig.~\ref{fig:intro}).
As the reward function is based on fundamental physical properties such as energy, this formulation is not restricted to the generation of molecules of a particular type.
We thus encourage the agent to implicitly learn the laws of atomic interaction from scratch to build molecules that go beyond what can be represented with graph-based RL methods.
To enable progress towards designing such molecules, we introduce a new RL environment called \textsc{MolGym}.
It comprises a suite of tasks in which both single molecules and molecule clusters need to be constructed.
For all of these tasks, we provide baselines using quantum-chemical calculations.
Finally, we propose a novel policy network architecture that 
can efficiently learn to solve these tasks by working in a translation and rotation invariant state-action space.

In summary, our contributions are as follows:
\begin{itemize}
    \item we propose a novel RL formulation for general molecular design in Cartesian coordinates (Section~\ref{sec:setup});
    \item we design a reward function based on the electronic energy, which we approximate via fast quantum-chemical calculations (Section~\ref{sec:reward});
    \item we present a translation and rotation invariant policy network architecture for molecular design (Section~\ref{sec:agent});
    \item we introduce \mbox{\textsc{MolGym}}, an RL environment comprising several molecular design tasks along with baselines based on quantum-chemical calculations (Section~\ref{sec:tasks});
    \item we perform experiments to evaluate the performance of our proposed policy network using standard policy gradient methods (Section~\ref{sec:results}).
\end{itemize}

\section{Reinforcement Learning for Molecular Design Guided by Quantum Mechanics}

In this section, we provide a brief introduction to RL and present our novel RL formulation for molecular design in Cartesian coordinates.

\subsection{Background: Reinforcement Learning}

In the standard RL framework, an agent interacts with the environment in order to maximize some reward.
We consider a fully observable environment with deterministic dynamics.
Such an environment is formally described by a Markov decision process (MDP) $\mathcal{M} = (\mathcal{S}, \mathcal{A}, \mathcal{T}, \mu_0, \gamma, T, r)$ with state space $\mathcal{S}$,
action space $\mathcal{A}$, transition function $\mathcal{T}: \mathcal{S} \times \mathcal{A} \mapsto S$, initial state distribution $\mu_0$, discount factor $\gamma \in (0, 1]$, time horizon $T$ and reward function $r: \mathcal{S} \times \mathcal{A} \mapsto \mathbb{R}$. The value function $V^\pi(s_t)$ is defined as the expected discounted return when starting from state $s_t$ and following policy $\pi$ thereafter, i.e. $V^\pi(s_t) = \mathbb{E}_{\pi} [ \sum_{t'=t}^T \gamma^{t'} r(s_{t'}, a_{t'}) \vert s_t ]$. The goal is to learn a stochastic policy $\pi(a_t \vert s_t)$ that maximizes the expected discounted return $J(\theta) = \mathbb{E}_{s_0 \sim \mu_0}[V^\pi(s_0)]$.

\textbf{Policy Gradient Algorithms} \quad
Policy gradient methods are well-suited for RL in continuous action spaces.
These methods learn a parametrized policy $\pi_\theta$ by performing gradient ascent in order to maximize $J(\theta)$.
More recent algorithms \citep{schulman2015trust, schulman2017proximal} improve the stability during learning by constraining the policy updates.
For example, proximal policy optimization (PPO) \citep{schulman2017proximal} employs a clipped surrogate objective.
Denoting the probability ratio between the updated and the old policy as $r_t(\theta) = \frac{\pi_\theta(a_t \vert s_t)}{\pi_{\theta_{\text{old}}}(a_t \vert s_t)}$,
the clipped objective $J^\text{CL}$ is given by
\begin{equation*}
    J^\text{CL}(\theta) = \mathbb{E} \left[ \min(r_t(\theta)\hat{A}_t, \text{clip}(r_t(\theta), 1-\epsilon, 1+\epsilon)\hat{A}_t) \right],
\end{equation*}
where $\hat{A}_t$ is an estimator of the advantage function, and $\epsilon$ is a hyperparameter that controls the interval beyond which $r(\theta)$ gets clipped.
To further reduce the variance of the gradient estimator, actor-critic approaches \citep{konda2000actor} are often employed.
The idea is to use the value function (i.e. the critic) to assist learning the policy (i.e. the actor).
If the actor and critic share parameters, the objective becomes
\begin{equation*}
    J^\text{AC}(\theta) = \mathbb{E} \left[J^\text{CL}(\theta) - c_1 J^\text{V} + c_2 \mathbb{H}[\pi_\theta \vert s_t] \right],
\end{equation*}
where $c_1$, $c_2$ are coefficients, $J^\text{V} = (V^\pi_\phi(s_t) - V^\text{target})^2$ is a squared-error loss, and $\mathbb{H}$ is an entropy regularization term to encourage sufficient exploration.

\subsection{Setup}\label{sec:setup}

\begin{figure}[t]
\centering
\includegraphics{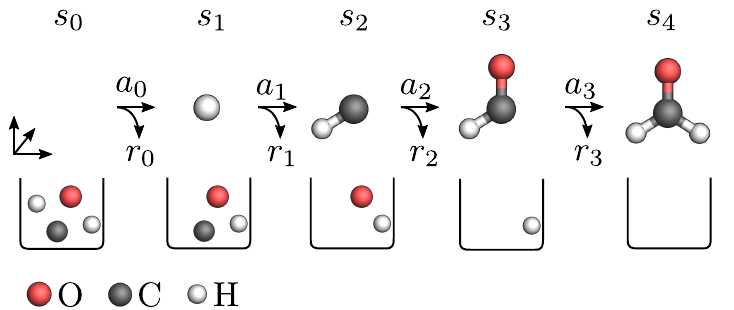}
\caption{
Rollout of an episode with bag $\beta_0 = \ce{CH2O}$.
The agent constructs a molecule by sequentially placing atoms from the bag onto the 3D canvas until the bag is empty.
}
\label{fig:setup}
\end{figure}

We design molecules by sequentially drawing atoms from a given bag and placing them onto a 3D \emph{canvas}.
This task can be formulated as a sequential decision-making problem in an MDP with deterministic transition dynamics, where
\begin{itemize}
    \item state $s_t = (\mathcal{C}_t, \beta_t)$ contains the canvas $\mathcal{C}_t = \mathcal{C}_0 \cup \{ (e_i, x_i) \}_{i=0}^{t-1}$,
    i.e. a set of atoms with chemical element $e_i \in \{\ce{H}, \ce{C}, \ce{N}, \ce{O}, \dots \}$
    and position $x_i \in \mathcal{R}^3$ placed until time $t-1$,
    as well as a bag $\beta_t = \left\{ (e, m(e)) \right\}$ of atoms still to be placed; $\mathcal{C}_0$ can either be empty, $\mathcal{C}_0 = \emptyset$, or contain a set of atoms,
    i.e. $\mathcal{C}_0 = \{ (e_i, x_i) \}$ for some $i \in \mathbb{Z}^-$; $m(e)$ is the multiplicity of the element $e$;
    \item action $a_t = (e_t, x_t)$ contains the element $e_t \in \beta_t$ and position $x_t \in \mathbb{R}^3$ of the next atom to be placed;
    \item deterministic transition function $\mathcal{T}(s_t, a_t)$ places an atom through action $a_t$ in state $s_t$, returning the next state $s_{t+1} = (\mathcal{C}_{t+1}, \beta_{t+1})$ with $\beta_{t+1} = \beta_t \backslash e_t$;
    \item reward function $r(s_t, a_t)$ quantifies how applying action $a_t$ in state $s_t$ alters properties of the molecule,
    e.g. the stability of the molecule as measured in terms of its quantum-chemical energy.
\end{itemize}

Fig.~\ref{fig:setup} depicts the rollout of an episode.
The initial state $(\mathcal{C}_0, \beta_0) \sim \mu_0(s_0)$ of the episode comprises the initial content $\mathcal{C}_0$ of the canvas and a bag of atoms $\beta_0$ to be placed,
e.g. $\mathcal{C}_0 = \emptyset$, and $\beta_0 = \ce{CH2O}$%
\footnote{Short hand for $\{(\ce{C}, 2), (\ce{H}, 2), (\ce{O}, 1) \}$.}
sampled uniformly from a given set of bags.
The agent then sequentially draws atoms from the bag without replacement and places them onto the canvas until the bag is empty.

\subsection{Reward Function}\label{sec:reward}
One advantage of designing molecules in Cartesian coordinates is that we can evaluate states in terms of quantum-mechanical properties,
such as the energy or dipole moment.
In this paper, we focus on designing \emph{stable} molecules, i.e. molecules with low energy $E \in \mathbb{R}$;
however, linear combinations of multiple desirable properties are possible as well (see Section~\ref{sec:tasks} for an example).
We define the reward $r(s_t, a_t) = -\Delta_E(s_t, a_t)$ as the negative difference in energy between the resulting molecule described by $\mathcal{C}_{t+1}$
and the sum of energies of the current molecule $\mathcal{C}_t$ and a new atom of element $e_t$,
i.e.
\begin{equation}\label{eq:energy}
    \Delta_E(s_t, a_t) = E(\mathcal{C}_{t+1}) - \left[E(\mathcal{C}_t) + E(e_t)\right],
\end{equation}
where $E(e) \coloneqq E(\{ (e, [0,0,0]\tran \})$.
Intuitively, the agent is rewarded for placing atoms so that the energy of the resulting molecules is low.
Importantly, with this formulation the undiscounted return for building a molecule is independent of the order in which atoms are placed.
If the reward only consisted of $E(C_{t+1})$, one would double-count interatomic interactions.
As a result, the formulation in Eq.~\eqref{eq:energy} prevents the agent from learning to greedily choose atoms of high atomic number first, as they have low intrinsic energy.

Quantum-chemical methods, such as the ones based on density functional theory (DFT), can be employed to compute the energy $E$ for a given $\mathcal{C}$.
Since such methods are computationally demanding in general,
we instead choose to evaluate the energy using the semi-empirical Parametrized Method 6 (PM6) \citep{Stewart2007}
as implemented in the software package \textsc{Sparrow} \citep{Husch2018a,Bosia2019}; see the Appendix for details.
PM6 is significantly faster than more accurate methods based on DFT and sufficiently accurate for the scope of this study.
For example, the energy $E$ of systems containing 10 atoms can be computed within hundreds of milliseconds with PM6;
with DFT, this would take minutes.
We note that more accurate methods can be used as well if the computational budget is available.

\section{Policy}\label{sec:agent}

Building molecules in Cartesian coordinates allows to 1) extend molecular design through deep RL to a much broader class of molecules compared to graph-based approaches, and 2) employ reward functions based on fundamental physical properties such as the energy.
However, working directly in Cartesian coordinates introduces several additional challenges for policy learning.

Firstly, it would be highly inefficient to naively learn to place atoms directly in Cartesian coordinates since molecular properties are invariant under symmetry operations such as translation and rotation.
For instance, the energy of a molecule---and thus the reward---does not change if the molecule gets rotated, 
yet an agent that is not taking this into account would need to learn these solutions separately.
Therefore, we require an agent that is \emph{covariant} to translation and rotation, i.e., if the canvas is rotated or translated, the position $x_t$ of the atom to be placed should be rotated and translated as well.
To achieve this, our agent first models the atom's position in \emph{internal} coordinates which are \emph{invariant} under translation and rotation.
Then, by mapping from internal to Cartesian coordinates, we obtain a position $x_t$ that features the required \emph{covariance}.
The agent's internal representations for states and actions are introduced in Sections~\ref{sec:state-rep} and \ref{sec:actor}, respectively.

Secondly, the action space contains both discrete (i.e. element $e_t$) and continuous actions (i.e. position $x_t$).
This is in contrast to most RL algorithms, which assume that the action space is either discrete or continuous.
Due to the continuous actions, policy exploration becomes much more challenging compared to graph-based approaches.
Further, not all discrete actions are valid in every state, e.g. the element $e_t$ has to be contained in the bag $\beta_t$.
These issues are addressed in Section~\ref{sec:actor}, where we propose a novel actor-critic neural network architecture for efficiently constructing molecules in Cartesian coordinates.

\subsection{State Representation}\label{sec:state-rep}

Given that our agent models the position of the atom to be placed in internal coordinates,
we require a representation for each atom on the canvas $\mathcal{C}$ that is invariant under translation and rotation of the canvas.%
\footnote{We omit the time index when it is clear from the context.}
To achieve this, we employ \textsc{SchNet} \citep{Schutt2017,schutt2018schnet}, a deep learning architecture consisting of continuous-filter convolutional layers that works directly on atoms placed in Cartesian coordinates.
$\text{SchNet}(\mathcal{C})$ produces an embedding of each atom in $\mathcal{C}$ that captures information about its local atomic environment.
As shown in Fig.~\ref{fig:model} (left), we combine this embedding $\tilde{\mathcal{C}}$ with a latent representation $\tilde{\beta}$ of the bag, yielding a \emph{state embedding} $\tilde{s}$, i.e.
\begin{equation}
    \tilde{s} = [\tilde{\mathcal{C}}, \tilde{\beta}],\qquad
    \tilde{\mathcal{C}} = \text{SchNet}(\mathcal{C}), \ \ \tilde{\beta} = \text{MLP}_\beta(\beta),
\end{equation}
where $\text{MLP}_\beta$ is a multi-layer perceptron (MLP).

\subsection{Actor}\label{sec:actor}

\begin{figure}[t]
\centering
\includegraphics{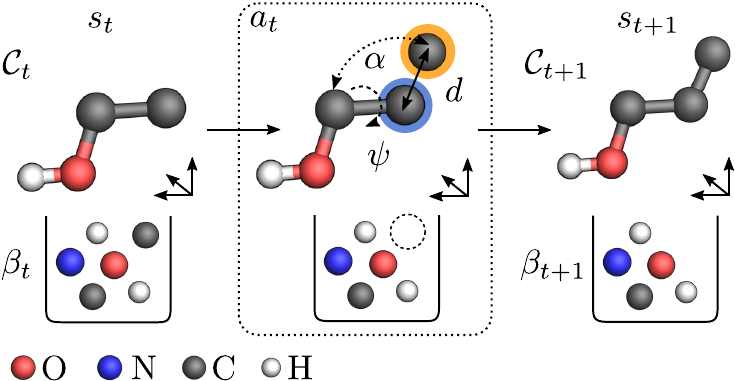}
\caption{
Construction of a molecule using an action-space representation that is invariant under translation and rotation.
\textbf{Left:} Current state $s_t$ with canvas $\mathcal{C}_t$ and remaining bag $\beta_t$.
\textbf{Center:} Action $a_t$ adds an atom from the bag (highlighted in orange) relative to the focus $f$ (highlighted in blue). The relative coordinates ($d$, $\alpha$, $\psi$) uniquely determine its absolute position.
\textbf{Right:} Resulting state $s_{t+1}$ after applying action $a_t$ in state $s_t$.
}
\label{fig:agent}
\end{figure}

\textbf{Action Representation} \quad%
We model the position of the atom to be placed in \emph{internal} coordinates---a commonly used representation for molecular structures in computational chemistry---relative to previously placed atoms.
If the canvas is initially empty, $\mathcal{C}_0 = \emptyset$, the agent selects an element $e_0$ from $\beta_0$ and places it at the origin, i.e. $a_0=(e_0, [0,0,0]\tran)$.
Once the canvas $\mathcal{C}_t$ contains at least one atom, the agent first decides on a focal atom, $f \in \{1, \dots, n(\mathcal{C}_t) \}$, where $n(\mathcal{C}_t)$ denotes the number of atoms in $\mathcal{C}_t$. This focal atom represents a local reference point close to which the next atom is going to be placed (see Fig.~\ref{fig:agent}).
The agent then models the position $x \in \mathbb{R}^3$ with respect to $f$ in internal coordinates $(d, \alpha, \psi)$, where
\begin{itemize}
    \item $d \in \mathbb{R}$ is the Euclidean distance between $x$ and the position $x_f$ of the focal atom;
    \item $\alpha \in [0, \pi]$ is the angle between the two lines defined by $(x, x_f)$ and $(x, x_{n1})$, where $x_{n1}$ is the position of the atom closest to $f$; if less than two atoms are on the canvas, $\alpha$ is undefined/unused.
    \item $\psi \in [-\pi, \pi]$ is the dihedral angle between two intersecting planes spanned by ($x$, $x_f$, $x_{n1}$) and ($x_f$, $x_{n1}$, $x_{n2}$), where $x_{n2}$ is the atom that is the second\footnote{In the unlikely event that two atoms are exactly equally far from the focal atom, a random order for $x_{n1}$ and $x_{n2}$ is chosen.} closest to the focal atom; if less than three atoms are on the canvas, $\psi$ is undefined/unused.
\end{itemize}
As shown in Fig.~\ref{sec:agent} (right), these internal coordinates can then be mapped back to Cartesian coordinates $x$.

\begin{figure*}[t]
\centering
\includegraphics[width=\textwidth]{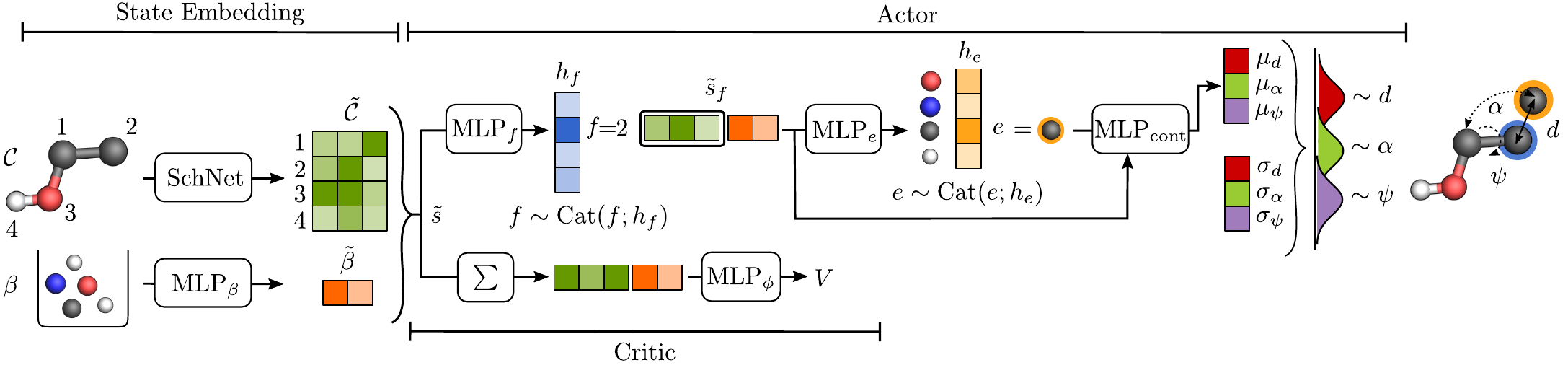}
\caption{
Illustration of the state embedding, actor and critic network.
The canvas $\mathcal{C}$ and the bag of atoms $\beta$ are fed to the state embedding network to obtain a translation and rotation invariant
state representation $\tilde{s}$.
The actor network then selects
1) a focal atom $f$,
2) an element $e$,
and 3) internal coordinates $(d, \alpha, \psi)$.
The critic takes the bag and the sum across all atoms on the canvas to compute a value $V$.
}
\label{fig:model}
\end{figure*}

\textbf{Model} \quad%
This action representation suggests a natural generative process:
first choose next to which focal atom the new atom is placed,
then select its element,
and finally decide where to place the atom relative to the focal atom.
Therefore, we assume that the policy factorizes as
\begin{align}
\pi_{\theta}(\psi, \alpha, d, e, f \vert s) =
&\ p(\psi, \alpha, d \vert e, f, s) \nonumber \\
\times &\ p(e \vert f, s) p(f \vert s).
\end{align}

We model the distributions over $f$ and $e$ as categorical, $\text{Cat}(h)$,
where $h_f \in \mathbb{R}^{n(\mathcal{C})}$ and $h_e \in \mathbb{R}^{E_\text{max}}$ are the logits predicted by separate MLPs,
and $E_\text{max}$ is the largest atomic number that can be selected.
Further, $p(\psi, \alpha, d \vert e, f, s)$ is factored into a product of univariate Gaussian distributions $\mathcal{N}(\mu, \sigma^2)$,
where the means $\mu_d$, $\mu_\alpha$ and $\mu_\psi$ are given by an MLP
and the standard deviations $\sigma_d$, $\sigma_\alpha$ and $\sigma_\psi$ are global parameters.
Formally,
\begin{align}
    h_f &= \text{MLP}_f(\tilde{s}), \\
    h_e &= \text{MLP}_e(\tilde{s}_f), \\
    \mu_d, \mu_\alpha, \mu_\psi  &= \text{MLP}_{\text{cont}}(\tilde{s}_f, \mathds{1}(e)),
\end{align}
where
$\tilde{s}_f = [\tilde{\mathcal{C}}_f, \tilde{\beta}]$ is the state embedding of the focal atom $f \sim \text{Cat}(f; h_f)$,
$\mathds{1}(e)$ is a one-hot vector representation of element $e \sim \text{Cat}(e; h_e)$,
and $d \sim \mathcal{N}(d; \mu_d, \sigma^2_d)$, $\alpha \sim \mathcal{N}(\alpha; \mu_\alpha, \sigma^2_\alpha)$, and $\psi \sim \mathcal{N}(\psi; \mu_\psi, \sigma^2_\psi)$
are sampled from their respective distributions.
The model is shown in Fig.~\ref{fig:model}. 

\textbf{Maintaining Valid Actions} \quad
As the agent places atoms onto the canvas during a rollout, the number of possible focal atoms $f$ increases and the number of elements $e$ to choose from decreases.
To guarantee that the agent only chooses valid actions, i.e. $f \in \{1, \dots, n(\mathcal{C})\}$ and $e \in \beta$, we mask out invalid focal atoms and elements by setting their probabilities to zero and re-normalizing the categorical distributions.
Neither the agent nor the environment makes use of ad-hoc concepts like valence or bond connectivity---any atom on the canvas can potentially be chosen.

\textbf{Learning the Dihedral Angle} \quad
The sign of the dihedral angle $\psi$ depends on the two nearest neighbors of the focal atom and is difficult to learn, especially if the two atoms are nearly equally close to the focal atom.
In practice, we therefore learn the absolute value $|\psi| \in [0, \pi]$ instead of $\psi$, as well as the sign $\kappa \in \{+1, -1\}$, such that $\psi = \kappa \vert\psi\vert$.
To estimate $\kappa$, we exploit the fact that the transition dynamics are deterministic.
We generate embeddings of both possible next states (for $\kappa = +1$ and $ \kappa = -1$) and select the embedding of the atom just added,
which we denote by $\tilde{s}_+$ and $\tilde{s}_-$.
We then choose $\kappa = +1$ over $\kappa = -1$ with probability%
\begin{equation}
    p_+ = \frac{\exp(u_+)}{\exp(u_+) + \exp(u_-)},
\end{equation}
such that $p(\kappa \vert \, \vert\psi\vert, \alpha, d, e, f, s) = \text{Ber}(\kappa; p_+)$,
where $u_\pm = \text{MLP}_\kappa(\tilde{s}_{\pm, })$;
we further motivate this choice in the Appendix.
Thus, the policy is given by $\pi_\theta(\kappa \vert\psi\vert, \alpha, d, e, f \vert s) = p(\kappa \vert \, \vert\psi\vert, \alpha, d, e, f, s) p(|\psi|, \alpha, d \vert e, f, s) p(e \vert f, s) p(f \vert s)$.

\subsection{Critic}

The critic needs to compute a value for the entire state $s$.
Since the canvas is growing as more atoms are taken from the bag and placed onto the canvas, a pooling operation is required.
Here, we compute the sum over all atomic embeddings $\tilde{C}_i$.
Thus, the critic is given by
\begin{equation}
    V_\phi(s) = \text{MLP}_\phi \left(\sum_{i=1}^{n(\mathcal{C})} \tilde{C}_i, \tilde{\beta} \right),
\end{equation}
where $\text{MLP}_\phi$ is an MLP that computes value $V$ (see Fig.~\ref{fig:model}).

\subsection{Optimization}
We employ PPO \citep{schulman2017proximal} to learn the parameters $(\theta, \phi)$ of the actor $\pi_\theta$ and critic $V_\phi$, respectively.
While most RL algorithms can only deal with either continuous or discrete action spaces and thus require additional modifications to handle both \citep{masson2016reinforcement,wei2018hierarchical,xiong2018parametrized},
PPO can be applied directly as is.
To help maintain sufficient exploration throughout learning, we include an entropy regularization term over the policy.
However, note that the entropies of the continuous and categorical distributions often have different magnitudes;
further, in this setting the entropies over the categorical distributions vary significantly throughout a rollout: as the agent places more atoms, the support of the distribution over valid focal atoms $f$ increases and the support of the distribution over valid elements $e$ decreases.
To mitigate this issue, we only apply entropy regularization to the categorical distributions, which we find to be sufficient in practice.

\section{Related Work}

\textbf{Deep Generative Models} \quad
A prevalent strategy for molecular design based on machine learning is to employ deep generative models.
These approaches first learn a latent representation of the molecules and then perform a search in latent space (e.g., through gradient descent) to discover new molecules with sought chemical properties.
For example, \citet{Gomez-Bombarelli2016,Kusner2017,Blaschke2018,Lim2018,Dai2018} utilized VAEs to perform search or optimization in a latent space to find new molecules.
\citet{Segler2018} used RNNs to design molecular libraries.
The aforementioned approaches generate SMILES strings, a linear string notation, to describe molecules \citep{Weininger1988}.
Further, there exist a plethora of generative models that work with graph representations of molecules (e.g., \citet{Jin2017,Bradshaw2018,Li2018a,Li2018b,Liu2018,DeCao2018,Bradshaw2019}).
In these methods, atoms and bonds are represented by nodes and edges, respectively.
\citet{Brown2019} developed a benchmark suite for graph-based generative models, showing that generative models outperform classical approaches for molecular design.
While the generated molecules are shown to be valid \citep{DeCao2018,Liu2018} and synthesizable \citep{Bradshaw2019}, the generative model is restricted to a (small) region of chemical space for which the graph representation is valid, e.g. single organic molecules.

\textbf{3D Point Cloud Generation} \quad
Another downside of string- and graph-based approaches is their neglect of information encoded in the interatomic distances.
To this end, \citet{Gebauer2018,Gebauer2019} proposed a generative neural network for sequentially placing atoms in Cartesian coordinates.
While their model respects local symmetries by construction, atoms are placed on a 3D grid.
Further, similar to aforementioned approaches, this model depends on a dataset to exist that covers the particular class of molecules for which one seeks to generate new molecules.

\textbf{Reinforcement Learning} \quad
\citet{Olivecrona2017}, \citet{Guimaraes2018}, \citet{Putin2018}, \citet{Neil2018} and \citet{Popova2018} presented RL approaches based on string representations of molecules.
They successfully generated molecules with given desirable properties but, similar to other generative models using SMILES strings, struggled with chemical validity.
\citet{You2018} proposed a graph convolutional policy network based on graph representations of molecules, where the reward function is based on empirical properties such as the drug-likeliness. While this approach was able to consistently produce valid molecules, its performance still depends on a dataset required for pre-training.
Considering the large diversity of chemical structures, the generation of a dataset that covers the whole chemical space is hopeless.
To address this limitation, \citet{Zhou2019} proposed an agent that learned to generate molecules from scratch using a Deep Q-Network (DQN) \citep{Mnih2015}.
However, such graph-based RL approaches are still restricted to the generation of single organic molecules for which this representation was originally designed.
Further, graph representations prohibit the use of reward functions based on fundamental physical laws, and one has to resort to heuristics instead.
Finally, geometric constraints cannot be imposed on the design process.
\citet{Jorgensen2019} introduced an atomistic structure learning algorithm, called \emph{ALSA}, that utilizes a convolutional neural network to build 2D structures and planar compounds atom by atom.

\begin{figure*}[t]
    \centering
    \includegraphics{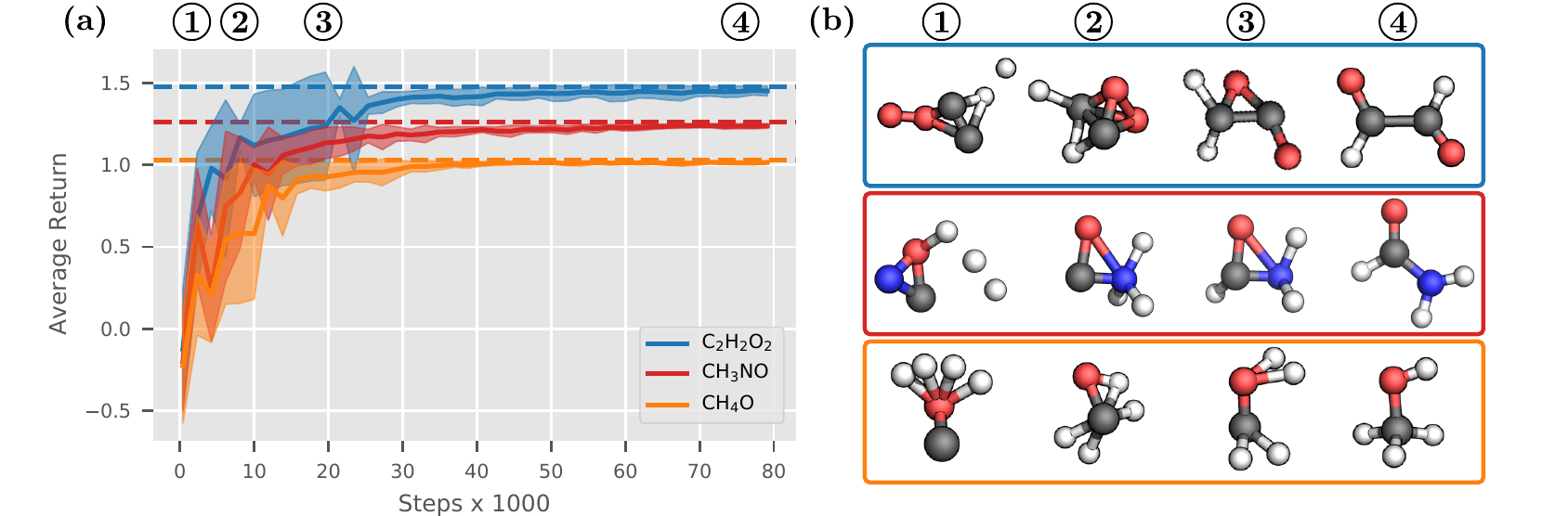}
    \caption{
    \textbf{(a)} Average offline performance on the \emph{single-bag} task for bags $\ce{CH3NO}$, $\ce{CH4O}$ and $\ce{C2H2O2}$ across 10 seeds.
    Dashed lines denote optimal returns for each bag, respectively.
    Error bars show two standard deviations.
    \textbf{(b)} Generated molecular structures at different terminal states over time, showing the agent's learning progress.
    }
    \label{fig:toy}
\end{figure*}

\begin{figure}[t]
    \centering
    \includegraphics{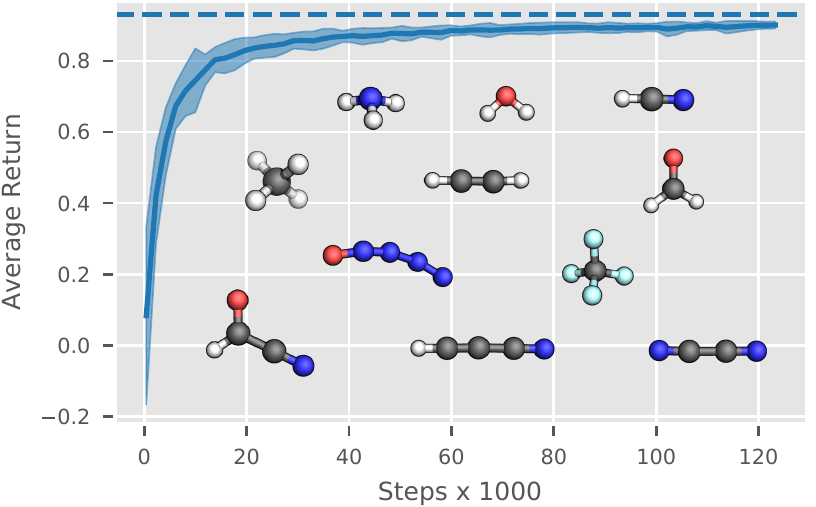}
    \caption{
    Average offline performance on the \emph{multi-bag} task, using 11 bags consisting of up to five atoms across 10 seeds.
    The dashed line denotes the optimal average return.
    Error bars show two standard deviations.
    The molecular structures shown are the terminal states at the end of training from one seed.
    }
    \label{fig:multi}
\end{figure}

\begin{figure*}[t]
    \centering
    \includegraphics{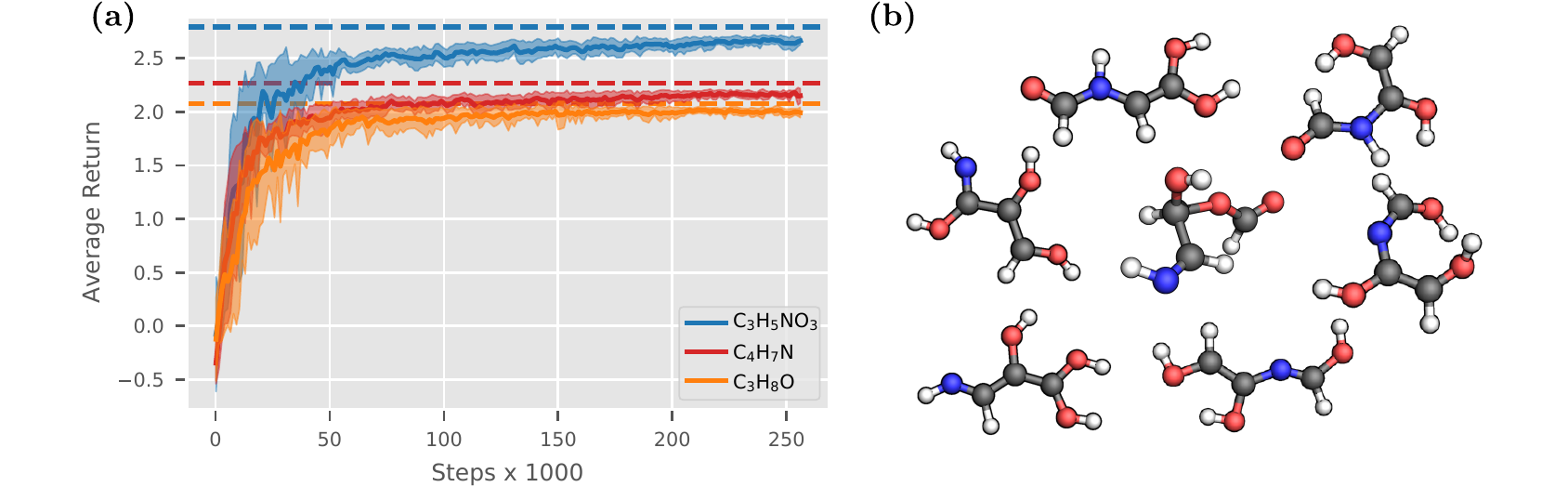}
    \caption{
    \textbf{(a)} Average offline performance on the \emph{single-bag} task for bags \ce{C3H5NO3}, \ce{C3H8O} and \ce{C4H7N} across 10 seeds.
    Dashed lines denote optimal return for each bag, respectively.
    Error bars show two standard deviations.
    \textbf{(b)} Selection of molecular structures generated by trained models for the bag \ce{C3H5NO3}.
    For the bags \ce{C3H8O} and \ce{C4H7N}, see the Appendix.
    }
    \label{fig:large}
\end{figure*}

\begin{figure}[t]
    \centering
    \includegraphics{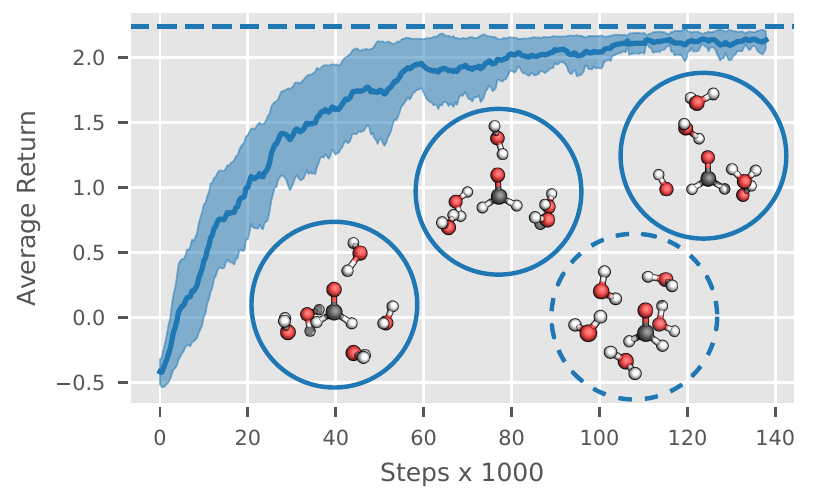}
    \caption{
    Average offline performance on the \emph{solvation} task with $5$ \ce{H2O} molecules across 10 seeds.
    Error bars show two standard errors.
    The plot is smoothed across five evaluations for better readability.
    The dashed line denotes the optimal return.
    A selection of molecular clusters generated by trained models are shown in solid circles;
    for comparison, a stable configuration obtained through structure optimization is depicted in a dashed circle.
    }
    \label{fig:solvation}
\end{figure}

\section{Experiments}\label{sec:experiments}

We perform experiments to evaluate the performance of the policy introduced in Section~\ref{sec:agent}.
While prior work has focused on building molecules using molecular graph representations,
we are interested in designing molecules in Cartesian coordinates. To this end, we introduce a new RL environment called \textsc{MolGym} in Section~\ref{sec:tasks}.
It comprises a set of molecular design tasks, for which we provide baselines using quantum-chemical calculations.
See the Appendix for details on how the baselines are determined.
\footnote{Source code of the agent and environment is available at \url{https://github.com/gncs/molgym}.}

We use \textsc{MolGym} to answer the following questions:
1) can our agent learn to construct single molecules in Cartesian coordinates from scratch,
2) does our approach allow building molecules across multiple bags simultaneously,
3) are we able to scale to larger molecules,
and 4) can our agent construct systems comprising multiple molecules?

\subsection{Tasks}\label{sec:tasks}

We propose three different tasks for molecular design in Cartesian coordinates, which are instances of the MDP formulation introduced in Section~\ref{sec:setup}:
\emph{single-bag}, \emph{multi-bag}, and \emph{solvation}.
More formally, the tasks are as follows:

\textbf{Single-bag} \quad
Given a bag, learn to design stable molecules.
This task assesses an agent's ability to build single stable molecules.
The reward function is given by $r(s_t, a_t) = -\Delta_E(s_t, a_t)$, see Eq.~\eqref{eq:energy}.
If the reward is below a threshold of $-0.6$, the molecule is deemed invalid and the episode terminates prematurely with the reward clipped at $-0.6$.%
\footnote{$\Delta_E$ is on the order of magnitude of $-0.1$~Hartree, resulting in a reward of around $0.25$ for a well placed atom.}

\textbf{Multi-bag} \quad
Given multiple bags with one of them being randomly selected before each episode, learn to design stable molecules.
This task focuses on the agent's capabilities to learn to build different molecules of different composition and size at the same time.
The same reward function as in the \emph{single-bag} task is used.
Offline performance is evaluated in terms of the average return across bags.
Similarly, the baseline is given by the average optimal return over all bags.

\textbf{Solvation} \quad
The task is to learn to place water molecules around an existing molecule (i.e. $\mathcal{C}_0$ is non-empty).
This task assesses an agent's ability to distinguish intra- and intermolecular interactions,
i.e. the atomic interactions within a molecule and those between molecules.
These interactions are paramount for the accurate description of chemistry in the liquid phase.
In this task, we deviate from the protocol used in the previous experiments as follows.
Initially, the agent is provided with an \ce{H2O} bag.
Once the bag is empty, the environment will refill it and the episode continues.
The episode terminates once $n \in \mathbb{N}^{+}$ bags of \ce{H2O} have been placed on the canvas.
By refilling the \ce{H2O} bag $n-1$ times instead of providing a single \ce{H_{2n}O_n} bag, the agent is guided towards building \ce{H2O} molecules.
\footnote{A comparison of the two protocols is given in the Appendix.}
The reward function is augmented with a penalty term for placing atoms far away from the center, i.e. $r(s_t, a_t) = -\Delta_E - \rho \|x\|_2$, where $\rho$ is a hyper-parameter.
This corresponds to a soft constraint on the radius at which the atoms should be placed.
This is a task a graph-based RL approach could not solve.

\subsection{Results}\label{sec:results}

In this section, we use the tasks specified in Section~\ref{sec:tasks} to evaluate our proposed policy.
We further assess the chemical validity, diversity and stability of the generated structures. Experiments were run on a 16-core Intel Xeon Skylake 6142 CPU with 2.6GHz and 96GB RAM. Details on the model architecture and hyperparameters are in the Appendix.

\textbf{Learning to Construct Single Molecules} \quad
In this toy experiment, we train the agent on the \emph{single-bag} task for the bags $\ce{CH3NO}$, $\ce{CH4O}$ and $\ce{C2H2O2}$, respectively.
Fig.~\ref{fig:toy} shows that the agent was able to learn the rules of chemical bonding and interatomic distances from scratch.
While on average the agent reaches $90\%$ of the optimal return after only $12\,000$ steps, the snapshots in Fig.~\ref{fig:toy} (b) highlight that the last $10\%$ determine chemical validity.
As shown in Fig.~\ref{fig:toy} (b), the model first learns the atomic distances $d$, followed by the angles $\alpha$ and the dihedral angles $\psi$.

\begin{table}[t]
\centering
\caption{
QM9 bags used in the experiments.
}
\label{tab:qm9}
\resizebox{\columnwidth}{!}{%
\begin{tabular}{@{}ll}
\toprule
Experiment & QM9 Bags Used \\ \midrule
Single-bag &  $\ce{C2H2O2}, \ce{CH3NO}, \ce{CH4O}$ \\
\multirow{2}{*}{Multi-bag} & $\ce{H2O}, \ce{CHN}, \ce{C2N2}, \ce{H3N}, \ce{C2H2}, \ce{CH2O}$, \\
          & $\ce{C2HNO}, \ce{N4O}, \ce{C3HN}, \ce{CH4}, \ce{CF4}$ \\
Single-bag (large) & $\ce{C3H5NO3}, \ce{C4H7N}, \ce{C3H8O}$ \\
\bottomrule
\end{tabular}
}
\end{table}

\textbf{Learning across Multiple Bags} \quad%
We train the agent on the \emph{multi-bag} task using all formulas contained in the QM9 dataset \citep{Ruddigkeit2012,Ramakrishnan2014} with up to $5$ atoms, resulting in $11$ bags (see Table~\ref{tab:qm9}).
Despite their small size, the molecules feature a diverse set of bonds (single, double, and triple) and geometries (linear, trigonal planar, and tetrahedral).
From the performance and from visual inspection of the generated molecular structures shown in Fig.~\ref{fig:multi},
it can be seen that a single policy is able to build different molecular structures across multiple bags.
For example, it learned that a carbon atom can have varying number and type of neighboring atoms leading to specific bond distance, angles, and dihedral angles.

\textbf{Scaling to Larger Molecules} \quad
To study our agent's ability to construct large molecules we let it solve the \emph{single-bag} task with the bags $\ce{C3H5NO3}$, $\ce{C3H8O}$, and $\ce{C4H7N}$.
Results are shown in Fig.~\ref{fig:large}.
After $154\,000$ steps,
the agent achieved an average return of
$2.60$ on \ce{C3H5NO3} (maximum across seeds at $2.72$, optimum at $2.79$),
$2.17$ on \ce{C4H7N} ($2.21$, $2.27$),
and $1.98$ on \ce{C3H8O} ($2.04$, $2.07$).
While the agent did not always find the most stable configurations, it was able to explore a diverse set of chemically valid structures (including bimolecular structures, see Appendix).

\textbf{Constructing Molecular Clusters} \quad
We task the agent to place $5$ water molecules around a formaldehyde molecule, i.e. $\mathcal{C}_0 = \ce{CH2O}$ and $n=5$.
The distance penalty parameter $\rho$ is set to $0.01$.%
\footnote{Further experiments on the \emph{solvation} task are in the Appendix.}
From Fig.~\ref{fig:solvation}, we observe that the agent is able to learn to construct \ce{H2O} molecules
and place them in the vicinity of the solute.
A good placement also allows for hydrogen bonds to be formed between water molecules themselves and between water molecules and the solute (see Fig.~\ref{fig:solvation}, dashed circle).
In most cases, our agent arranges \ce{H2O} molecules such that these bonds can be formed (see Fig.~\ref{fig:solvation}, solid circles).
The lack of hydrogen bonds in some structures could be attributed to the approximate nature of the quantum-chemical method used in the reward function.
Overall, this experiment showcases that our agent is able to learn both intra- and intermolecular interactions, going beyond what graph-based agents can learn.


\begin{table}[ht]
\centering
\caption{Assessment of generated structures in different experiments by chemical validity, RMSD (in \AA), and diversity.}
\label{tab:struct_assess}
\resizebox{\columnwidth}{!}{%
\begin{tabular}{llrrr}
\toprule
Task & Experiment & Validity & RMSD & Diversity \\
\midrule
\multirow{3}{*}{Single-bag}
    & \ce{C2H2O2} & 0.90 & 0.32 & 3 \\
    & \ce{CH3NO}  & 0.70 & 0.20 & 3 \\
    & \ce{CH4O}   & 0.80 & 0.11 & 1 \\
\midrule
\multirow{1}{*}{Multi-bag}
    & -    & 0.78 & 0.05 & 22 \\
\midrule
\multirow{4}{*}{\shortstack[l]{Single-bag\\(large)}}
    & \ce{C3H5NO3}  & 0.70  & 0.39 & 40 \\
    & \ce{C4H7N}    & 0.80  & 0.29 & 20 \\
    & \ce{C3H8O}    & 0.90  & 0.47 &  4 \\
    & \ce{C7H8N2O2}  & 0.60 & 0.61 & 61 \\
\midrule
\multirow{3}{*}{Solvation}
   & \ce{Formaldehyde}  & 0.80 & 1.03 & 1 \\ 
   & \ce{Acetonitrile}  & 0.90 & 1.06 & 1 \\ 
   & \ce{Ethanol}       & 0.90 & 0.92 & 1 \\
\bottomrule
\end{tabular}
}
\end{table}

\textbf{Quality Assessment of Generated Molecules} \quad In the spirit of the \emph{GuacaMol} benchmark \citep{Brown2019}, we assess the molecular structures generated by the agent with respect to chemical validity, diversity and structural stability for each experiment. 
To enable a comparison with existing approaches, we additionally ran experiments with the bag \ce{C7H8N2O2}, the stoichiometry of which is taken from the \emph{GuacaMol} benchmark \citep{Brown2019}.

The results are shown in Table~\ref{tab:struct_assess}. To determine the validity and stability of the generated structures, we first took the terminal states of the last iteration for a particular experiment.
Structures are considered valid if they can be successfully parsed by \textsc{RDKit} \citep{RDKit2019093}.
However, those consisting of multiple molecules 
were not considered valid (except in the \emph{solvation} task).
The validity reported in Table~\ref{tab:struct_assess} is the ratio of valid molecules over 10 seeds.

All valid generated structures underwent a structure optimization using the PM6 method (see Appendix for more details). 
Then, the RMSD (in \AA) between the original and the optimized structure were computed.
In Table~\ref{tab:struct_assess}, the median RMSD over all generated structures is given per experiment.
In the approach by \citet{Gebauer2019}, an average RMSD of $\approx$ 0.25~\AA{} is reported.
Due to significant differences in approach, application, and training procedure we forego a direct comparison of the methods.

Further, two molecules are considered identical if the SMILES strings generated by \textsc{RDKit} are the same.
The diversity reported in Table~\ref{tab:struct_assess} is the total number of unique and valid structures generated through training over 10 seeds.

\section{Discussion}

This work is a first step towards general molecular design through RL in Cartesian coordinates.
One limitation of the current formulation is that we need to provide bags for which we know good solutions exist when placed completely.
While being able to provide such prior knowledge can be beneficial, we are currently restricted to designing molecules of known formulas.
A possible solution is to provide bags that are larger than necessary, e.g. generated randomly or according to some fixed budget for each element, and enable the agent to stop before the bag is empty.

Compared to graph-based approaches, constructing molecules by sequentially placing atoms in Cartesian coordinates greatly increases the flexibility in terms of the type of molecular structures that can be built.
However, it also makes the exploration problem more challenging: whereas in graph-based approaches a molecule can be expanded by adding a node and an edge,
here, the agent has to learn to precisely position an atom in Cartesian coordinates from scratch.
As a result, the molecules we generate are still considerably smaller.
Several approaches exist to mitigate the exploration problem and improve scalability, including:
1) \emph{hierarchical RL}, where molecular fragments or entire molecules are used as high-level actions;
2) \emph{imitation learning}, in which known molecules are converted into expert trajectories;
and 3) \emph{curriculum learning}, where the complexity of the molecules to be built increases over time.

\section{Conclusion}

We have presented a novel RL formulation for molecular design in Cartesian coordinates, in which the reward function is based on quantum-mechanical properties such as the energy.
We further proposed an actor-critic neural network architecture based on a translation and rotation invariant state-action representation.
Finally, we demonstrated that our model can efficiently solve a range of molecular design tasks from our \textsc{MolGym} RL environment from scratch.

In future work, we plan to increase the scalability of our approach and enable the agent to stop before a given bag is empty.
Moreover, we are interested in combining the reward with other properties such as drug-likeliness and applying our approach to other classes of molecules, e.g. transition-metal catalysts.

\section*{Acknowledgements}
We would like to thank the anonymous reviewers for their valuable feedback. We further thank Austin Tripp and Vincent Stimper for useful discussions and feedback.
GNCS acknowledges funding through an Early Postdoc.Mobility fellowship by the Swiss National Science Foundation (P2EZP2\_181616).
RP receives funding from iCASE grant \#1950384 with support from Nokia.


\bibliography{references}
\bibliographystyle{icml2020}



\newpage
\clearpage
\appendix

\section{Quantum-Chemical Calculations}\label{app:quantum}

For the calculation of the energy $E$ we use the fast semi-empirical Parametrized Method 6 (PM6) \citep{Stewart2007}.
In particular, we use the implementation from the software package \textsc{Sparrow} \citep{Husch2018a,Bosia2019}.
For each calculation, a molecular charge of zero and the lowest possible spin multiplicity are chosen.
All calculations are spin-unrestricted. 

Limitations of semi-empirical methods are highlighted in, for example, recent work by \citet{Husch2018b}.
More accurate methods such as approximate density functionals need to be employed especially for systems containing transition metals.

For the quantum-chemical calculations to converge reliably, we ensured that atoms are not placed too close ($<$ 0.6~\AA)
nor too far away from each other ($>$ 2.0~\AA).
If the agent places an atom outside these boundaries, the minimum reward of $-0.6$ is awarded and the episode terminates.

\section{Learning the Dihedral Angle}\label{app:preference}

\begin{figure}[ht]
    \centering
    \includegraphics{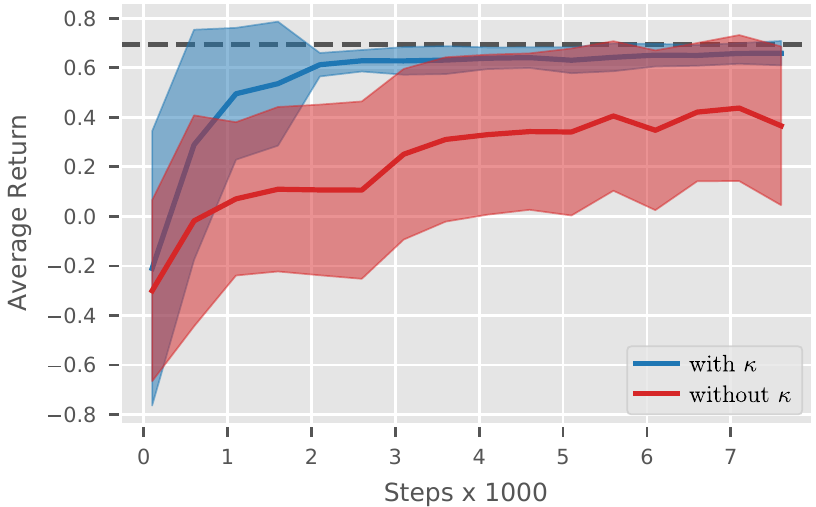}
    \caption{
    Average offline performance on the \emph{single-bag} task for the bag \ce{CH4} across 10 seeds. 
    Estimating $\kappa$ and $|\psi|$ separately (with $\kappa$) significantly speeds up learning compared to estimating $\psi$ directly (without $\kappa$).
    Error bars show two standard deviations. 
    The dashed line denotes the optimal return.
    }
    \label{fig:kappa}
\end{figure}

We experimentally validate the benefits of learning $|\psi| \in [0, \pi]$ and $\kappa \in \{-1, 1\}$ instead of $\psi \in [-\pi, \pi]$ by comparing the two models on the \emph{single-bag} task with bag \ce{CH4} (methane).
Methane is one of the simplest molecules that requires the model to learn a dihedral angle.
As shown in Fig.~\ref{fig:kappa}, learning the sign of the dihedral angle separately (with $\kappa$) speeds up learning significantly.
In fact, the ablated model (without $\kappa$) fails to converge to the optimal return even after $100\,000$ steps (not shown).

\section{Experimental Details}\label{app:experiments}


\subsection{Model Architecture}

The model architecture is summarized in Table~\ref{tab:models}.
We initialize the biases of each MLP with $0$ and each weight matrix as a (semi-)orthogonal matrix.
After each hidden layer, a ReLU non-linearity is used.
The output activations are shown in Table~\ref{tab:models}.
As explained in the main text, both $\text{MLP}_f$ and $\text{MLP}_e$ use a masked softmax activation function to guarantee that only valid actions are chosen.
Further, we rescale the continuous actions $(\mu_d, \mu_\alpha, \mu_\psi) \in [-1, 1]^3$ predicted by $\text{MLP}_\text{cont}$ to ensure that $\mu_d \in [d_\text{min}, d_\text{max}]$, $\mu_\alpha \in [0, \pi]$ and $\mu_\psi \in [0, \pi]$.
For more details on the SchNet, see the original work \citep{schutt2018schnet}.

\begin{table}[ht]
\centering
\caption{Model architecture for actor and critic networks.}
\label{tab:models}
\resizebox{\columnwidth}{!}{%
\begin{tabular}{@{}lrc@{}}
\toprule
Operation & Dimensionality & Activation \\ \midrule
$\text{SchNet}$ & $n(\mathcal{C}) \times 4, *, n(\mathcal{C}) \times 64$ & $*$ (cf. Table~\ref{tab:hyperparams_schnet}) \\
$\text{MLP}_\beta$ & $e_{\text{max}}, 128, 32$ & linear \\
$\text{tile}$ & $32, n(\mathcal{C}) \times 32$ & --- \\
$\text{concat}$ & $n(\mathcal{C}) \times (64, 32), n(\mathcal{C}) \times 96$ & --- \\
$\text{MLP}_f$ & $n(\mathcal{C}) \times 96, n(\mathcal{C}) \times 128, n(\mathcal{C}) \times 1$ & softmax \\
$\text{select}$ & $n(\mathcal{C}) \times 96, 96$ & --- \\
$\text{MLP}_e$ & $96, 128, e_{\text{max}}$ & softmax \\
$\text{concat}$ & $(96, e_{\text{max}}), 96 + e_{\text{max}}$ & --- \\
$\text{MLP}_\text{cont}$ & $96 + e_{\text{max}}, 128, 3$ & tanh \\
$\text{MLP}_\kappa$ & $2 \times 96, 2 \times 128, 2 \times 1$ & softmax \\
$\text{pooling}$ & $n(\mathcal{C}) \times 96, 96$ & --- \\
$\text{MLP}_\phi$ & $96, 128, 128, 1$ & linear \\ \bottomrule
\end{tabular}
}
\end{table}

\subsection{Hyperparameters}

We manually performed an initial hyperparameter search on a single holdout validation seed.
The considered hyperparameters and the selected values are listed in Table~\ref{tab:hyperparams-singlebag} (\emph{single-bag}), Table~\ref{tab:hyperparams-multibag} (\emph{multi-bag}) and Table~\ref{tab:hyperparams-cluster} (\emph{solvation}).
The hyperparameters used for SchNet are shown in Table~\ref{tab:hyperparams_schnet}.

\begin{table}[ht]
\centering
\caption{Hyperparameters for the \emph{single-bag} task. Adapted values for the scalability (large) experiment are in parentheses.}
\label{tab:hyperparams-singlebag}
\resizebox{\columnwidth}{!}{%
\begin{tabular}{@{}lr@{}r@{}}
\toprule
Hyperparameter & Search set & ~~ Value (large) \\
\midrule
Range $[d_\text{min}, d_\text{max}]$ (\AA) & --- & ~~ $[0.95, 1.80]$ \\
Max. atomic number $e_\text{max}$ & --- & $10$ \\
Workers & --- & $16$ \\
Clipping $\epsilon$ & --- & $0.2$ \\
Gradient clipping & --- & $0.5$ \\
GAE parameter $\lambda$ & --- & $0.95$ \\
VF coefficient $c_1$ & --- & $1$ \\
Entropy coefficient $c_2$ & \{$0.00, 0.01$, $0.03$\} & $0.01$ \\
Training epochs & $\{5, 10\}$ & $5$ \\
Adam stepsize & $\{10^{-4},  3 \times 10^{-4} \}$ & $3 \times 10^{-4}$ \\
Discount $\gamma$ & $\{0.99, 1.00 \}$ & $0.99$ \\
Time horizon $T$ & $\{192, 256 \}$ & $192$ ($256$)  \\
Minibatch size & $\{24, 32 \}$ & $24$ ($32$) \\
\bottomrule
\end{tabular}
}
\end{table}

\begin{table}[ht]
\centering
\caption{Hyperparameters for the \emph{multi-bag} task.}
\label{tab:hyperparams-multibag}
\resizebox{\columnwidth}{!}{%
\begin{tabular}{@{}lr@{}r@{}}
\toprule
Hyperparameter & Search set & ~~ Value \\
\midrule
Range $[d_\text{min}, d_\text{max}]$ (\AA) & --- & ~~ $[0.95, 1.80]$ \\
Max. atomic number $e_\text{max}$ & --- & $10$ \\
Workers & --- & $16$ \\
Clipping $\epsilon$ & --- & $0.2$ \\
Gradient clipping & --- & $0.5$ \\
GAE parameter $\lambda$ & --- & $0.95$ \\
VF coefficient $c_1$ & --- & $1$ \\
Entropy coefficient $c_2$ & \{$0.00, 0.01$, $0.03$\} & $0.01$ \\
Training epochs & $\{5, 10\}$ & $5$ \\
Adam stepsize & $\{10^{-4},  3 \times 10^{-4} \}$ & $3 \times 10^{-4}$ \\
Discount $\gamma$ & $\{0.99, 1.00 \}$ & $0.99$ \\
Time horizon $T$ & $\{384, 512 \}$ & $384$  \\
Minibatch size & $\{48, 64 \}$ & $48$ \\
\bottomrule
\end{tabular}
}
\end{table}

\begin{table}[ht]
\centering
\caption{Hyperparameters for the \emph{solvation} task.}
\label{tab:hyperparams-cluster}
\resizebox{\columnwidth}{!}{%
\begin{tabular}{@{}lr@{}r@{}}
\toprule
Hyperparameter & Search set & ~~ Value \\
\midrule
Range $[d_\text{min}, d_\text{max}]$ (\AA) & --- & ~~ $[0.90, 2.80]$ \\
Max. atomic number $e_\text{max}$ & --- & $10$ \\
Distance penalty $\rho$ & --- & $0.01$ \\
Workers & --- & $16$ \\
Clipping $\epsilon$ & --- & $0.2$ \\
Gradient clipping & --- & $0.5$ \\
GAE parameter $\lambda$ & --- & $0.95$ \\
VF coefficient $c_1$ & --- & $1$ \\
Entropy coefficient $c_2$ & \{$0.00, 0.01$, $0.03$\} & $0.01$ \\
Training epochs & $\{5, 10\}$ & $5$ \\
Adam stepsize & $\{10^{-4},  3 \times 10^{-4} \}$ & $3 \times 10^{-4}$ \\
Discount $\gamma$ & $\{0.99, 1.00 \}$ & $0.99$ \\
Time horizon $T$ & $\{384, 512 \}$ & $384$  \\
Minibatch size & $\{48, 64 \}$ & $48$ \\
\bottomrule
\end{tabular}
}
\end{table}

\begin{table}[ht]
\centering
\caption{Hyperparameters for SchNet \citep{Schutt2018c} used in all experiments.}
\label{tab:hyperparams_schnet}
\begin{tabular}{@{}lr@{}r@{}r@{}}
\toprule
Hyperparameter & Search set & ~~Value   \\
\midrule
Number of interactions & --- & $3$ \\
Cutoff distance (\AA) & --- & $5.0$ \\
Number of filters & --- & $128$ \\
Number of atomic basis functions & $\{32, 64, 128 \}$ & $64$ \\
\bottomrule
\end{tabular}
\end{table}

\section{Baselines}\label{app:baselines}

Below, we report how the baselines for the \emph{single-bag} and \emph{multi-bag} tasks were derived.
First, we took all molecular structures for a given chemical formula (i.e. bag) from the QM9 dataset \citep{Ruddigkeit2012,Ramakrishnan2014}.
Subsequently, we performed a structure optimization using the PM6 method (as described in Section~\ref{app:quantum}) on the structures.
This was necessary as the structures in this dataset were optimized with a different quantum-chemical method.
Then, the most stable structure was selected and considered \emph{optimal} for this chemical formula; the remaining structures were discarded.
Since the undiscounted return is path independent, we determined the return $R(s)$ by computing the total interaction energy in the canvas $\mathcal{C}$, i.e.
\begin{equation}\label{eq:baseline}
    R(s) = E(\mathcal{C}) - \sum_{i=1}^{N} E(e_i),
\end{equation}
where $N$ is the number of atoms placed on the canvas.

The baseline for the \emph{solvation} task was determined in the following way.
12 molecular clusters were generated by randomly placing $n$ \ce{H2O} molecules around the solute molecule (in the main text $n=5$).
Subsequently, the structure of these clusters was optimized with the PM6 method (as described in Section~\ref{app:quantum}).
Similar to Eq.~\eqref{eq:baseline}, the undiscounted return of each cluster can be computed:
\begin{equation}
    R(s) = E(\mathcal{C}) - E(\mathcal{C}_0) - \sum_{i=1}^{N} \left\{ E(e_i) + \rho \| x_i \|_2 \right\},
\end{equation}
where the distance penalty $\rho = 0.01$.
Finally, the maximum return over the optimized clusters was determined.

\section{Additional Results}

\subsection{Single-bag Task}

\begin{figure}[ht]
\centering
\includegraphics{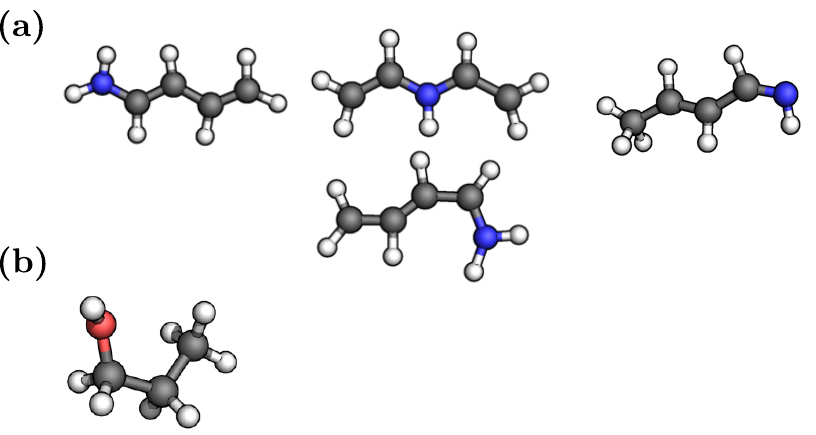}
\caption{
Selection of molecular structures generated by trained models for the bags \ce{C4H7N} (a) and \ce{C3H8O} (b).
}
\label{fig:large_app}
\end{figure}

\begin{figure}[ht]
\centering
\includegraphics[width=0.2\columnwidth]{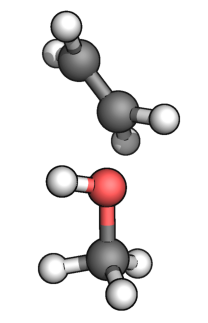}
\caption{
Bimolecular structure generated by a trained model for the bag \ce{C3H8O} in the \emph{single-bag} task.
}
\label{fig:bimolecular}
\end{figure}

\begin{figure}[ht]
\centering
\includegraphics{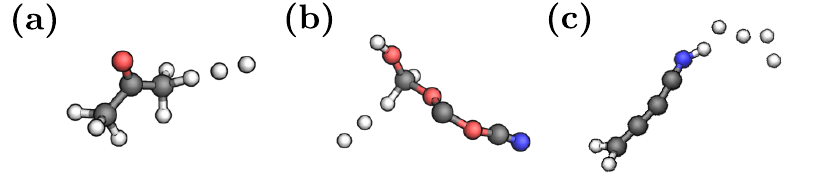}
\caption{
Selection of chemically invalid molecular structures generated by trained models for the bags \ce{C3H8O} (a), \ce{C3H5NO3} (b), and \ce{C4H7N} (c).
}
\label{fig:large_fail}
\end{figure}

In Fig.~\ref{fig:large_app}, we show a selection of molecular structures generated by trained models for the bags \ce{C4H7N} and \ce{C3H8O}.
Further, since the agent is agnostic to the concept of molecular bonds, it is able to build multiple molecules if it results in a higher return.
An example of a bimolecular structure generated by a trained model for the bag \ce{C3H8O} is shown in Fig.~\ref{fig:bimolecular}.
Finally, in Fig.~\ref{fig:large_fail}, we showcase a set of generated molecular structures that are not chemically valid.

\subsection{Solvation Task}

\begin{figure}[ht]
\centering
\includegraphics{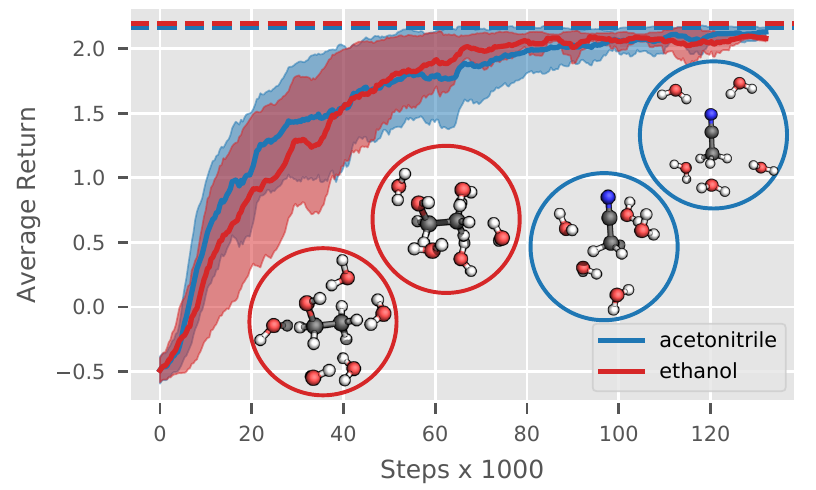}
\caption{
Average offline performances across 10 seeds on the \emph{solvation} task with $n=5$ and the initial states being acetonitrile and ethanol.
Error bars show two standard errors.
The plot is smoothed across five evaluations for better readability.
The dashed lines denote the optimal returns.
A selection of molecular clusters generated by trained models are shown in circles.
}
\label{fig:solvation_app}
\end{figure}

In Fig.~\ref{fig:solvation_app}, we report the average offline performances of agents placing 5 \ce{H2O} molecules around the solutes (i.e, $\mathcal{C}_0$) acetonitrile and ethanol.
As can be seen, the agents are able to accurately place water molecules such that they interact with the solute.
However, we stress that more accurate quantum-chemical methods for computing the reward are required to describe hydrogen bonds to chemical accuracy.

\begin{figure}[ht]
\centering
\includegraphics{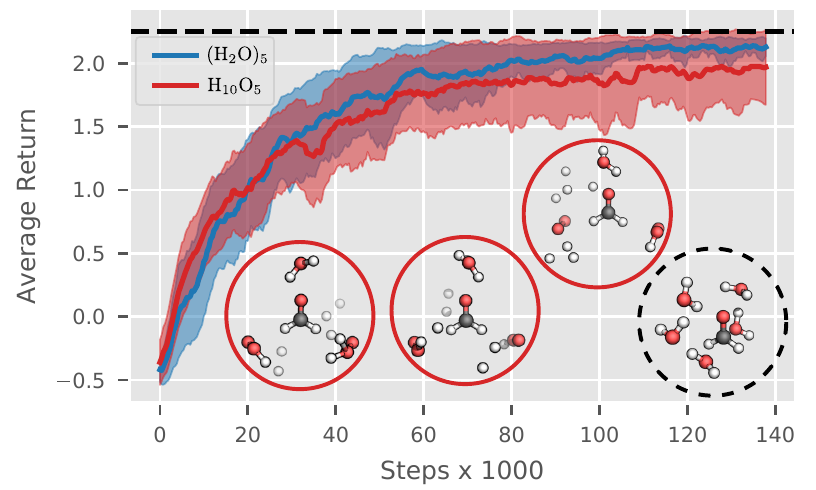}
\caption{
Average offline performance for the \emph{solvation} task with $n=5$ (blue) and placing atoms from a single \ce{H_{10}O_5} bag (red).
In both experiments, $\mathcal{C}_0$ is formaldehyde.
Error bars show two standard errors.
The plot is smoothed across five evaluations for better readability.
The dashed line denotes the optimal return.
A selection of molecular clusters generated by models trained on the \ce{H_{10}O_5} bag are shown in red solid circles;
for comparison, a stable configuration obtained through structure optimization is depicted in a black dashed circle.
}
\label{fig:solvation_comp}
\end{figure}

In Fig.~\ref{fig:solvation_comp}, we compare the average offline performance of two agents placing in total 10 \ce{H} and 5 \ce{O} atoms around a formaldehyde molecule.
One agent is given $5$ \ce{H2O} bags consecutively following the protocol of the \emph{solvation} task as described in the main text,
another is given a single \ce{H_{10}O_5} bag.
Their average offline performances are shown in Fig.~\ref{fig:solvation_comp} in blue and red, respectively.
It can be seen that giving the agent $5$ \ce{H2O} bags one at a time instead of a single \ce{H_{10}O_5} bag improves performance.

\subsection{Generalization and Transfer Learning}

\begin{figure}[ht]
\centering
\includegraphics{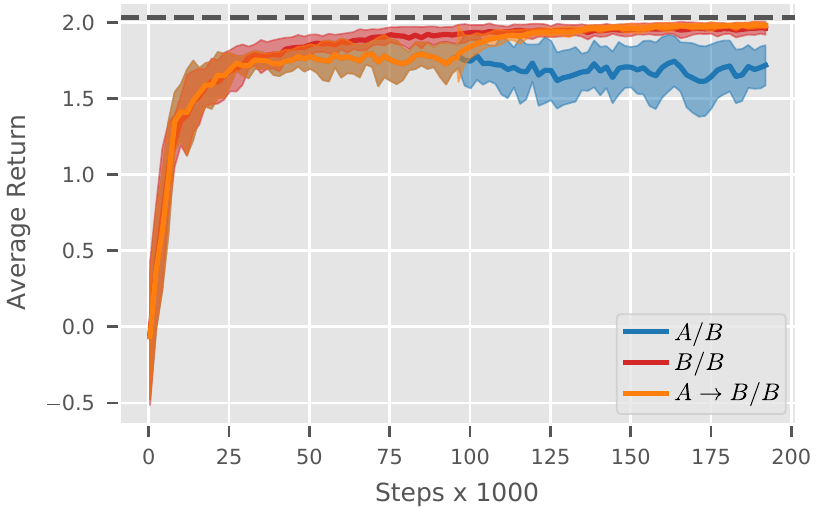}
\caption{
Average offline performance for agents 
$A/B$: trained on bags $A$ of size 6 and tested on bags $B$ of size 8, 
$B/B$: trained and tested on $B$, 
and $A \rightarrow B / B$: trained on $A$ for $96\,000$ steps, and fine-tune and tested on $B$.
See main text for more details. 
Error bars show two standard deviations.
The dashed line denotes the optimal average return.
}
\label{fig:generalization}
\end{figure}

To assess the generalization capabilities of our agent when faced with previously unseen bags, we train an agent on bags 
$A = \{ \ce{C2H2O2}, \ce{C2H3N}, \ce{C3H2O}, \ce{C3N2O}, \ce{CH3NO},\\ \ce{CH4O} \}$ of size 6 
and test on bags $B = \{ \ce{C3H2O3}, \ce{C3H4O},\\ \ce{C4H2O2}, \ce{CH4N2O}, \ce{C4N2O2}, \ce{C5H2O} \}$ of size 8.
As shown in Fig.~\ref{fig:generalization}, the agent $A / B$ achieves an average return of $1.79$, which is approximately $88\%$ of the optimal return.
In comparison, an agent trained and tested on $B$ ($B / B$) reaches an average return of $1.96$ (or $0.97\%$ of the optimal return).
We additionally train an agent on $A$ for $96\,000$ steps, and then fine-tune and test on $B$.
The agent $A \rightarrow B / B$ reaches the same performance as if trained from scratch within $20\,000$ steps of fine-tuning, showing successful transfer.
We anticipate that training on more bags and incorporating best practices from multi-task learning would further improve performance.


\end{document}